\pdfoutput=1

\documentclass[11pt]{article}

\usepackage[]{acl}

\usepackage{times}
\usepackage{latexsym}

\usepackage{graphicx}

\usepackage{amsmath}
\usepackage{amssymb}

\usepackage[ruled]{algorithm2e}

\usepackage[T1]{fontenc}

\usepackage[utf8]{inputenc}

\usepackage{microtype}

\newcommand{\drex}{\textsc{d-rex}}
\newcommand{\rank}{\textit{R}}
\newcommand{\expl}{\textit{EX}}
\newcommand{\rerank}{\textit{RR}}
\newcommand{\joint}{\textit{Joint}}

\title{\drex: Dialogue Relation Extraction with Explanations}

\author{Alon Albalak\textsuperscript{1}, Varun Embar\textsuperscript{2}, Yi-lin Tuan\textsuperscript{1}, Lise Getoor\textsuperscript{2}, William Yang Wang\textsuperscript{1} \\
  \textsuperscript{1}University of California, Santa Barbara\qquad \textsuperscript{2}University of California, Santa Cruz \\
  \texttt{\{alon\_albalak, ytuan\}@ucsb.edu} \\
  \texttt{\{vembar, getoor\}@ucsc.edu} \\
  \texttt{william@cs.ucsb.edu}
  }

\begin{document}
\maketitle

\begin{abstract}
Existing research studies on cross-sentence relation extraction in long-form multi-party conversations aim to improve relation extraction without considering the explainability of such methods.
This work addresses that gap by focusing on extracting explanations that indicate that a relation exists while using only partially labeled explanations.
We propose our model-agnostic framework, \drex, a policy-guided semi-supervised algorithm that optimizes for explanation quality and relation extraction simultaneously. We frame relation extraction as a re-ranking task and include relation- and entity-specific explanations as an intermediate step of the inference process. We find that human annotators are 4.2 times more likely to prefer D-REX's explanations over a joint relation extraction and explanation model. 
Finally, our evaluations show that \drex\;is simple yet effective and improves relation extraction performance of strong baseline models by 1.2-4.7\%.\footnote{Code and data publicly available at \url{https://github.com/alon-albalak/D-REX}}
\end{abstract}

\section{Introduction}

Traditional relation extraction (RE) approaches discover relations that exist between entities within a single sentence. Recently, several approaches have been proposed which focus on cross-sentence RE, the task of extracting relations between entities that appear in separate sentences \cite{Peng2017CrossSentenceNR,quirk-poon-2017-distant,Han2020AND,yao2019docred} as well as cross-sentence RE in dialogues \cite{yu-etal-2020-dialogue,Chen2020DialogueRE,Xue2021GDPNetRL,Qiu2021SocAoGIG,lee-choi-2021-graph}.

\begin{figure}
\centering
\includegraphics[width=0.95\columnwidth]{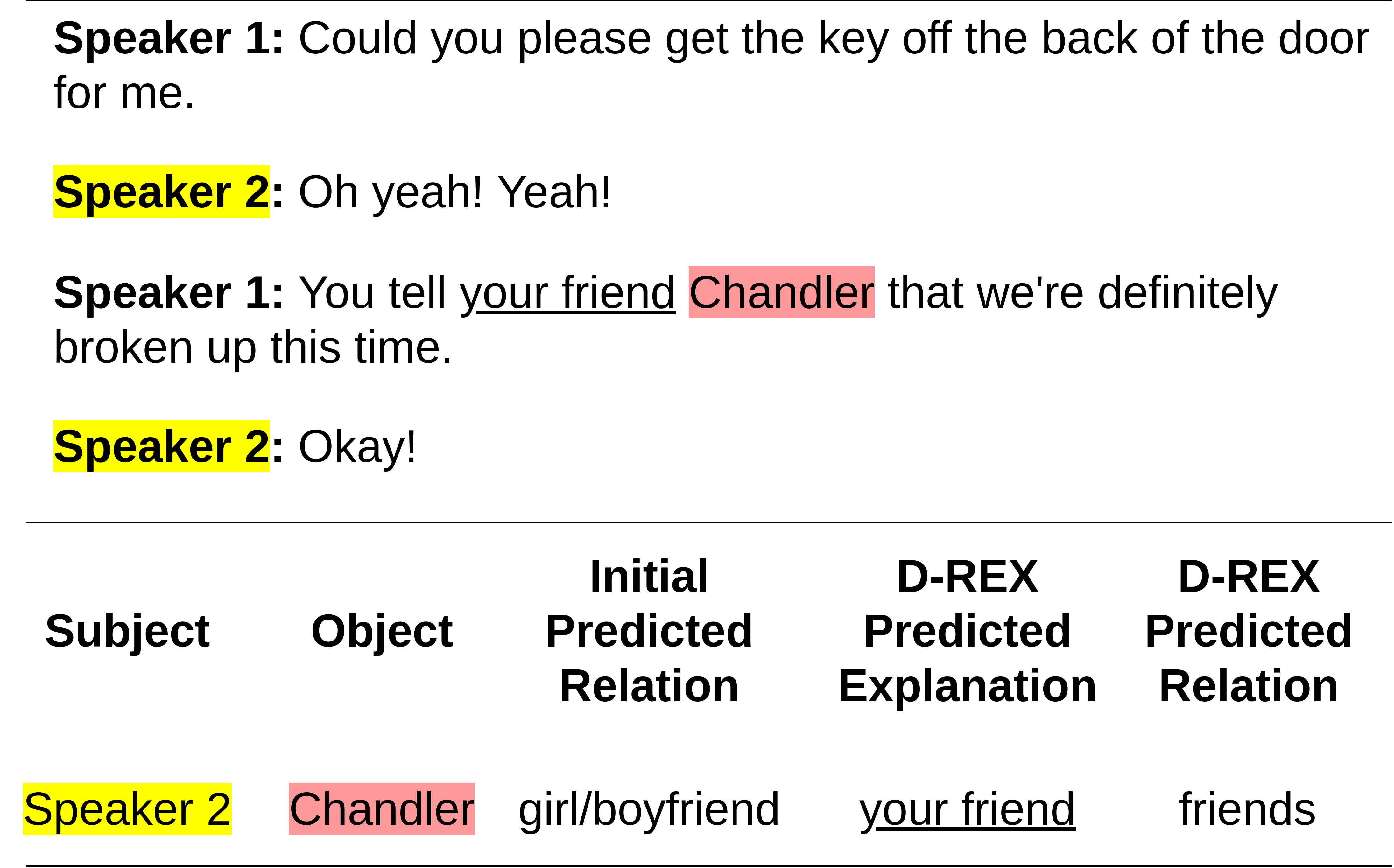} 
\caption{A sample dialogue between 2 speakers with actual \drex\;predictions. The model initially classifies Speaker 2 and chandler, incorrectly, as girl/boyfriend. After predicting the explanation "\underline{your friend}", \drex\;correctly re-ranks the relation as friends.}
\label{figure:1}
\end{figure}

A crucial step towards performing cross-sentence RE in multi-entity and multi-relation dialogues is to understand the context surrounding relations and entities (e.g., who said what, and to whom). Figure \ref{figure:1} shows an example from the DialogRE dataset where a simple BERT-based model (Initial Predicted Relation in Figure \ref{figure:1}) gets confused by multiple entities and relations existing in the same dialogue \cite{yu-etal-2020-dialogue}. The model predicts the ``girl/boyfriend'' relation between Speaker 2 and Chandler, however, it is clear from the context that the ``girl/boyfriend'' relation is referring to a different pair of entities: Speaker 1 and Chandler.

One approach to encourage a model to learn the context surrounding a relation is by requiring the model to generate an explanation along with the relation \cite{NEURIPS2018_4c7a167b}. In addition to the DialogRE dataset, \citet{yu-etal-2020-dialogue} introduces manually annotated \emph{trigger words} which they show play a critical role in dialogue-based RE. They define trigger words as ``the smallest span of contiguous text which clearly indicates the existence of the given relation''. In the context of RE, these trigger words can be used as potential explanations.

Our work aims to extract explanations that clearly indicate a relation while also benefiting an RE model by providing cross-sentence reasoning. Our proposed approach, \drex, makes use of multiple learning signals to train an explanation extraction model. First, \drex\;utilizes trigger words as a partial supervision signal.
Additionally, we propose multiple reward functions used with a policy gradient, allowing the model to explore the explanation space and find explanations that benefit the re-ranking model. Including these reward functions allows \drex\;to learn meaningful explanations on data with less than 40\% supervised triggers.

In order to predict relation- and entity-specific explanations in \drex, we pose RE as a relation re-ranking task with explanation extraction as an intermediate step and show that this is not possible for a model trained to perform both tasks jointly.

Our contributions are summarized as follows:
\begin{itemize}
    \item We propose \drex, \textbf{D}ialogue \textbf{R}elation \textbf{E}xtraction with e\textbf{X}planations, a novel system trained by policy gradient and semi-supervision.
    
    
    \item We show that \drex\;outperforms a strong baseline in explanation quality, with human evaluators preferring \drex\;explanations over 90\% of the time.
    
    \item We demonstrate that by conditioning on \drex\;extracted explanations, relation extraction models can improve by 1.2-4.7\%.
    
\end{itemize}

\section{Problem Formulation}

We follow the problem formulation of \citeauthor{yu-etal-2020-dialogue}: let \( d = (s_1 : u_1, s_2 : u_2, \dots , s_n : u_n) \) be a dialogue where $s_i$ and $u_i$ denote the speaker ID and the utterance from the $i^{th}$ turn, respectively. Let $\mathcal{E}, \mathcal{R}$ be the set of all entities in the dialogue and the set of all possible relations between entities, respectively. Each dialogue is associated with $m$ relational triples \textless\(s,r,o\)\textgreater\;where $s,o\in\mathcal{E}$ are subject and object entities in the given dialogue and $r\in\mathcal{R}$ is a relation held between the $s$ and $o$. Each relational triple may or may not be associated with a trigger $t$. It is important to note that there is no restriction on the number of relations held between an entity pair; however, there is at most one trigger associated with a relational triple. In this work, we consider an explanation to be of high quality if it strongly indicates that a relation holds, and for this purpose we consider triggers to be short explanations, though not always optimal in quality.

\subsection{Relation Extraction (RE)}
Given a dialogue $d$, subject $s$, and object $o$, the goal of RE is to predict the relation(s) that hold between $s$ and $o$. We also consider RE with additional evidence in the form of a trigger or predicted explanation. Formally, this is the same as relation extraction with an additional explanation, $ex$.

\subsection{Explanation Extraction (EE)}
We formulate EE as a span prediction problem. Given a dialogue $d$ consisting of $n$ tokens $T_{1}$ through $T_{n}$, and a relational triple \textless\(s,r,o\)\textgreater, the goal of EE is to predict start and end positions, $i,j$ in the dialogue, such that the explanation $ex=[T_{i},T_{i+1},\dots,T_{j}]$ indicates that $r$ holds between $s$ and $o$.
\section{Baseline Models}
\label{baseline_models}
We first introduce approaches for RE and EE based on state-of-the-art language models. We then propose a multitask approach that performs both tasks jointly. Our approaches use BERT\textsubscript{base} \cite{devlin-etal-2019-bert} and RoBERTa\textsubscript{base} \cite{roberta} pre-trained models\footnote{Pre-trained models obtained from https://github.com/huggingface/transformers \cite{wolf-etal-2020-transformers}}, and follow their respective fine-tuning protocols.

For all models, we maintain a single input format, which follows from \citeauthor{yu-etal-2020-dialogue}. Formally, for a dialogue $d$, subject $s$, object $o$, relation $r$, and explanation $ex$, the input sequence to all models is [CLS]\{$r/ex$[SEP]\}$s$[SEP]$o$[SEP]$d$, where \{$r/ex$[SEP]\} denotes that the relation or explanation may be included depending on the task setting. For RoBERTa models, we use the \textless s\textgreater\;and \textless/s\textgreater\;
tokens rather than [CLS] and [SEP], respectively.

\subsection{Relation Extraction (RE)}
\label{RE}
We follow the fine-tuning protocols of \citeauthor{devlin-etal-2019-bert} and \citeauthor{roberta} for BERT and RoBERTa classification models by using the output corresponding to the first token $C\in\mathbb{R}^{H}$ ([CLS] and \textless s\textgreater, respectively) as a latent representation of the entire input and train a classification matrix $W\in\mathbb{R}^{K \mathrm{x} H}$, where $K$ is the number of relation types and $H$ is the dimension of the output representations from the language model. For each relation $r_{i}$, the probability of $r_{i}$ holding between $s$ and $o$ in $d$ is calculated as 
$P_{i}=\mbox{sigmoid}(CW_{i}^{T})$. We compute the standard cross-entropy loss for each relation as
\begin{equation}\mathcal{L}_{RE}=-\dfrac{1}{\mathrm{K}}\sum_{i=1}^{\mathrm{K}}y_{i}\cdot \log (P_{i})+(1-y_{i})\cdot \log(1-P_{i})
\label{equation1}
\end{equation}
where $y_{i}$ denotes whether relation $i$ holds.


\begin{figure*}[t]
    \centering
    \includegraphics[scale=0.75]{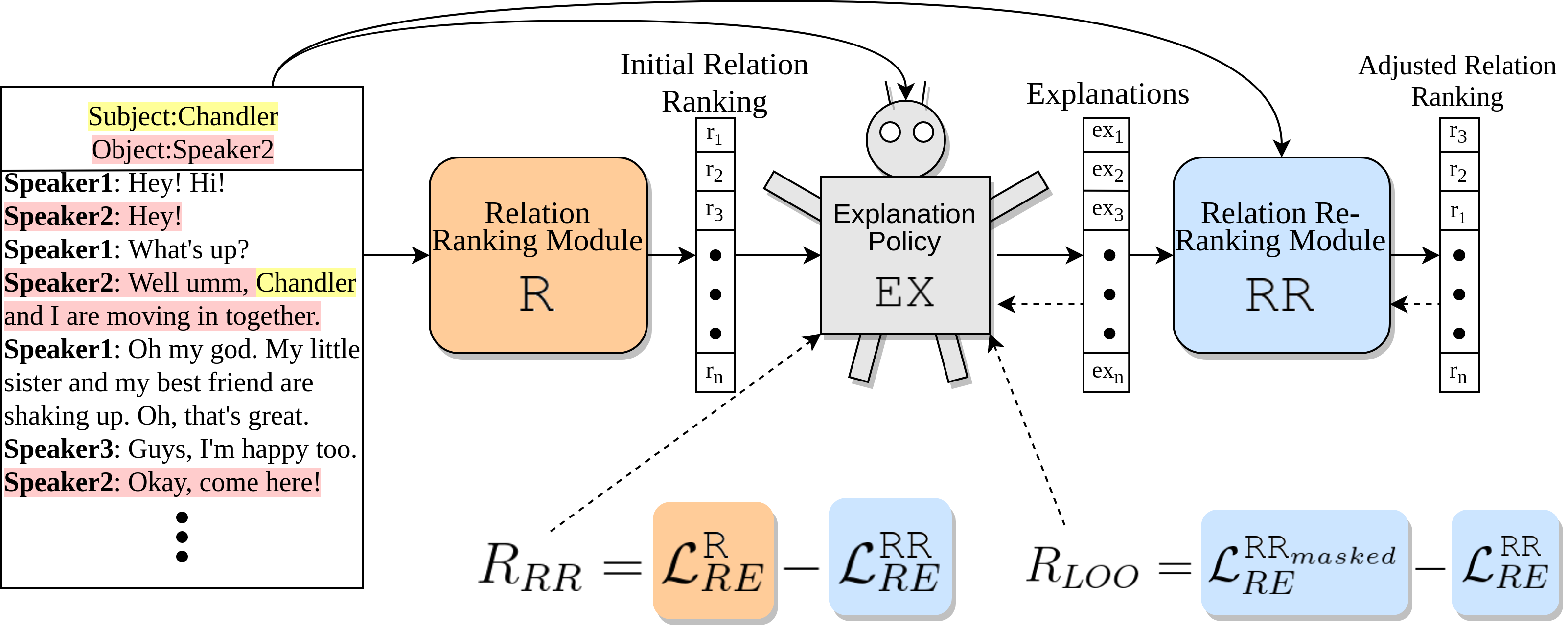}
    \caption{Overview of the \drex\;system. The relation $\boldsymbol{R}$anking module ranks relations conditioned only on the subject, object, and the dialogue. The $\boldsymbol{EX}$planation policy extracts supporting evidence for the ranked relations by conditioning on individual relations in addition to the original input. The relation $\boldsymbol{R}$e$\boldsymbol{R}$anking module conditions its rankings on supporting evidence from the explanation policy. In this hypothetical example, we see that relation 3 was originally ranked number 3 but had strong supporting evidence and was re-ranked in the number 1 spot. Solid lines represent model inputs/outputs, and dotted lines represent learning signals. Reward functions, $\mathcal{R}_{RR}$ and $\mathcal{R}_{LOO}$, are detailed in equations \ref{equation4} and \ref{equation5}, respectively.}
    \label{fig:training}
\end{figure*}

\subsection{Explanation Extraction (EE)}
\label{EE}
For EE, we use the input described above, with a natural language phrasing of a relation appended to the beginning of the sequence. For example, if $r$ is "per:positive\textunderscore impression", then we concatenate  "person positive impression" to the beginning.

We follow the fine-tuning protocol of \citeauthor{devlin-etal-2019-bert} for span prediction. We introduce start and end vectors, $S, E\in\mathbb{R}^{H}$. If $T_{i}\in\mathbb{R}^{H}$ is the final hidden representation of token $i$, then we compute the probability of token $i$ being the start of the predicted explanation as a dot product with the start vector, followed by a softmax over all words in the dialogue:
\begin{equation}
    P_{T_{i}}^{S}=\dfrac{exp(S\cdot T_{i})}{\sum_{j}{exp(S\cdot T_{j})}}
\label{equation2}
\end{equation}
To predict the end token, we use the same formula and replace the start vector $S$ with the end vector $E$. To compute the loss, we take the mean of the cross-entropy losses per token for the start and end vectors. Formally, let $|d|$ be the number of tokens in dialogue $d$, then
\begin{equation}
\begin{split}\mathcal{L}_{EX}&=-\dfrac{1}{|d|}\sum_{i}^{|d|}\\
&\big( y_{i}^{S}\cdot log(P_{T_{i}}^{S})+(1-y_{i}^{S})\cdot\log (1-P_{T_{i}}^{S})\big)\\
+&\big( y_{i}^{E}\cdot log(P_{T_{i}}^{E}) + (1-y_{i}^{E})\cdot\log(1-P_{T_{i}}^{E}) \big)
\label{equation3}
\end{split}
\end{equation}
where $y_{i}^{S}$ and $y_{i}^{E}$ are the start and end labels.
Because we want explanations extracted only from the dialogue, if the start or end token with largest log-likelihood occurs within the first $l$ tokens, where $l$ is the length of [CLS]$r$[SEP]$s$[SEP]$o$[SEP], then we consider there to be no predicted explanation.

\subsection{Joint Relation and Explanation Model}
\label{sec:joint}
The joint RE and EE model uses the standard input from $\S$\ref{baseline_models}. It utilizes a BERT or RoBERTa backbone, and has classification and span prediction layers identical to those in the RE and EE models. Similarly, the loss is computed as the weighted sum of RE and EE losses:
\[\mathcal{L_{J}} = \alpha\mathcal{L}_{RE}+(1-\alpha)\mathcal{L}_{EX}
\]
where $\alpha$ is an adjustable weight. In practice, we find that $\alpha=0.5$ works best.
\paragraph{Flaw of the joint model}
The disadvantage of the joint model is this: supposing that an entity pair has 2 relations, each explanation should be paired with a single relation. However, by making predictions jointly, there is no guaranteed mapping from predicted explanations to predicted relations. One method of solving this issue is to predict relations and explanations in separate steps. It is possible to first predict relations and then condition the explanation prediction on each individual relation and conversely. This idea forms the basis for \drex.
\section{\drex}
\label{sec:drex}
In this section, we introduce the \drex\;system. 
We begin by introducing the models which make up the system. Next, we present the training and inference algorithms. Finally, we discuss the optimization objectives for each model in the system.

\subsection{Models}
The \drex\;framework requires three components: an initial relation ranking model, an explanation model, and a relation re-ranking model, shown in Figure \ref{fig:training}.

\paragraph{Initial Ranking Model (\rank)}
In our algorithm and discussions, we use \rank\;to denote the initial ranking model. There are no restrictions on \rank, it can be any algorithm which ranks relations (e.g., deep neural network, rule-based, etc.) such as \cite{yu-etal-2020-dialogue,lee-choi-2021-graph}. However, if \rank\;needs to be trained, it must be done prior to \drex\;training; \drex\;will not make any updates to \rank.

In our evaluations, we use the relation extraction model described in $\S$\ref{RE}. The input to this model is ($s$,$o$,$d$) and the output is a ranking, \rank$(s,o,d)$.

\paragraph{Explanation Extraction Model (\expl)}
In our algorithm and discussions, we use \expl\;to denote the explanation model. In this paper we limit our experiments to extractive explanation methods, as opposed to generative explanation methods, however this is not a limitation of \drex. The only limitation on the explanation model is that we require it to produce human-interpretable explanations. Thus, it is also possible to use generative models such as GPT-2 \cite{radford2019language} or graph-based methods such as \cite{yu-ji-2016-unsupervised,Xue2021GDPNetRL} with adjustments to the formulation of the reward functions.

In our evaluations, we use the model as described in $\S$\ref{EE}. The input to \expl\;is ($r$,$s$,$o$,$d$) and the output is an extracted phrase from $d$, denoted as \expl$(r,s,o,d)$.

\paragraph{Relation Re-Ranking Model (\rerank)}
\label{rerank}
In our algorithm and discussions, we let \rerank\;denote the relation re-ranking model. In the \drex\;training algorithm, \rerank\;is updated through gradient-based optimization methods, and must be able to condition its ranking on explanations produced by \expl. In our experiments, we use the same model architecture as \rank\;and include an explanation as additional input to the model. The input to \rerank\;is ($ex$,$s$,$o$,$d$) and the output is a relation ranking, denoted as \rerank$(ex,s,o,d)$.

\subsection{\drex\;Algorithm}
The outline of this algorithm is shown in pseudocode in Algorithm \ref{algorithm1}.

Assuming that we have ranking, explanation, and re-ranking models \rank, \expl, \rerank, then given a single datum ($s,r,o,t,d$), comprised of a subject, relation, object, trigger(may be empty), and dialogue, the \drex\;algorithm operates as follows:
The ranking model takes as input ($s,o,d$) and computes the probability of each relation from the predefined relation types. Next, we take the top-k ranked relations, $r_{pred}=\rank(s,o,d)_{1:k}$, and compute explanations. For $i=1,...,k$, explanations are computed as $ex_{i}=\expl(r_{pred_{i}},s,o,d)$. Finally, for each predicted explanation, the re-ranking model computes $k$ probabilities for each relation type, using $(ex_{i},s,o,d)$ as the input to \rerank. The final probabilities for each relation type are computed as the mean across all $k$+1 predictions from \rank\;and \rerank.

\begin{algorithm}[t]
\small
\SetKwInOut{Input}{Input}

\Input{Pre-trained ranking, explanation, and re-ranking models: \rank, \expl, \rerank\newline k: for number of relations to re-rank}
\caption{The proposed training algorithm for \drex}
\KwData{Dataset: $\mathcal{D}$}
\For{($s,r,o$,$t$,$d$) in $\mathcal{D}$}{
Compute ranking loss: $\mathcal{L}_{RE}^{\rank}(s,o,d)$

$r_{pred}\leftarrow$ \rank(s,o,d)$_{1:k}$

\For{
i in $r_{pred}$}{
$ex_{i}\leftarrow$ \expl($r_{pred_{i}},s,o,d$)

Compute Re-ranking loss: $\mathcal{L}_{RE}^{\rerank}(ex_{i},s,o,d)$
\tcp*{Equation \ref{equation1}}

Compute Re-Ranking Reward: $\mathcal{R}_{RR}$
\tcp*{Equation \ref{equation4}}

Compute Leave-one-out Reward: $\mathcal{R}_{LOO}$
\tcp*{Equation \ref{equation5}}

Compute policy gradient with rewards $R_{RR},R_{LOO}$
\tcp*{Equation \ref{equation6}}
}
\If{$t$ not empty}{
Compute $\mathcal{L}_{EX}$
\tcp*{Equation \ref{equation3}}
}
Update \expl,\rerank parameters with calculated losses
}
\label{algorithm1}
\end{algorithm}

\subsection{Model optimization}
We propose multiple optimization objectives to train an \expl\;model that extracts explanations meaningful to humans and beneficial to the relation extraction performance while ensuring that \rerank\;maintains high-quality predictions.

\paragraph{Explanation Model Optimization}
We train \expl\;with supervision on labeled samples, and a policy gradient for both labeled and unlabeled samples, allowing for semi-supervision. For the policy gradient, we introduce two reward functions: a relation re-ranking reward and a leave-one-out reward.

\textbf{Re-ranking Reward} The purpose of the re-ranking reward is to ensure that \expl\;predicts explanations which benefit \rerank. Formally, let $\mathcal{L}_{RE}^{\rank}(s,o,d)$ be the loss for \rank, given the subject, object, and dialogue: $s,o,d$. And let $\mathcal{L}_{RE}^{\rerank}(ex,s,o,d)$ be the loss of \rerank, given the explanation, subject, object, and dialogue: $ex,s,o,d$. Then we define the relation re-ranking reward as: \begin{equation}
\mathcal{R}_{RR}=\mathcal{L}_{RE}^{\rank}(s,o,d)-\mathcal{L}_{RE}^{\rerank}(ex,s,o,d)
\label{equation4}
\end{equation}
Because \rank\;is stationary, \expl\;maximizes this function by minimizing $\mathcal{L}_{RE}^{\rerank}$. Of course, \expl\;can only minimize $\mathcal{L}_{RE}^{\rerank}$ through its predicted explanations.

\textbf{Leave-one-out Reward} The purpose of the leave-one-out reward is to direct \expl\;in finding phrases which are essential to correctly classifying the relation between an entity-pair. This reward function is inspired by previous works which make use of the leave-one-out idea for various explanation purposes \cite{shahbazi-etal-2020-relation,DBLP:journals/corr/LiMJ16a}. We can calculate the leave-one-out reward using either \rank\;or \rerank, and it is calculated by finding the difference between the standard relation extraction loss and the loss when an explanation has been masked. Formally, if $d$ is the original dialogue and $ex$ is the given explanation, let $d_{mask}(ex)$ be the dialogue with $ex$ replaced by mask tokens. Then, the leave-one-out reward is defined as: \begin{equation}
\mathcal{R}_{LOO}=\mathcal{L}_{RE}(s,o,d_{mask}(ex))-\mathcal{L}_{RE}(s,o,d)
\label{equation5}
\end{equation}
Because $\mathcal{L}_{RE}$ is calculated using the same model for both the masked and unmasked loss, \expl\;maximizes this reward function by maximizing the masked loss. Of course, the only interaction that \expl\;has with the masked loss is through the explanation it predicts.

\textbf{Policy Gradient} We view \expl\;as an agent whose action space is the set of all continuous spans from the dialogue. In this view, the agent interacts with the environment by selecting two tokens, a start and end token and receives feedback in the form of the previously discussed reward functions. Let $i,j$ be the start and end indices that the explanation model selects and $T_{i}$ be the $i^{th}$ token, then $ex=d[i:j]=[T_{i},T_{i+1},\dots,T_{j}]$ and the probabilities of $i,j$ being predicted are calculated as $P_{T_{i}}^{S}$ and $P_{T_{j}}^{E}$ according to equation \ref{equation2}.

For both reward functions, we use a policy gradient \cite{sutton2018reinforcement} to update the weights of the explanation model and calculate the loss as\begin{equation}
    \mathcal{L}_{EX_{PG}}=-(log(P_{T_{i}}^{S})+log(P_{T_{j}}^{E}))*(R_{RR}+R_{LOO})
    \label{equation6}
\end{equation}

Additionally, while training \expl\;in the \drex\;algorithm, we make use of supervision when available. In the case where supervision exists, we calculate an additional loss, $\mathcal{L}_{EX}$, as defined in equation \ref{equation3}.

\paragraph{Relation Extraction Re-ranking Model Optimization}
While training \drex\; we train \rerank\;with labeled relations as supervision and use a cross-entropy loss, $\mathcal{L}_{RE}^{\rerank}$, calculated in the same way as \rank\;in Equation \ref{equation1}.
\section{Experimental Evaluation}

In this section, we present an evaluation of \drex\;in comparison with baselines methods on the relation extraction and explanation extraction tasks.

\subsection{Experimental settings}
For our experiments, we re-implement the BERT\textsubscript{S} model from \cite{yu-etal-2020-dialogue} as well as a new version which replaces BERT with RoBERTa. In our paper, we refer to these models as \rank\textsubscript{BERT} and \rank\textsubscript{RoBERTa}.
All models are implemented in PyTorch\footnote{https://pytorch.org/} and Transformers\cite{wolf-etal-2020-transformers}, trained using the AdamW optimizer \cite{loshchilov2018fixing}. All experiments were repeated five times and we report mean scores along with standard deviations. \drex\;models use a top-k of five and are initialized from the best performing models with the same backbone. For example, \drex\textsubscript{BERT} uses two copies of \rank\textsubscript{BERT} \cite{yu-etal-2020-dialogue} to initialize the ranking and re-ranking models and \expl\textsubscript{BERT} to initialize the explanation model. When training \joint, we do not calculate $\mathcal{L}_{EX}$ for relational triples without a labeled trigger. The full details of our training settings are provided in Appendix \ref{hyperparameters}.

\begin{table}[t]
\small
\begin{center}
    \begin{tabular}{c|c|c|c}
    \multicolumn{4}{c}{\textbf{DialogRE V2}}\\
    \hline
    \begin{tabular}{c}
    Dial-\\ogues\end{tabular} & 
    \begin{tabular}{c}Rela-\\tions
    \end{tabular} & \begin{tabular}{c}
    Relational \\ Triples \\ (train/dev/ \\ test)
    \end{tabular}
    & 
    \begin{tabular}{c} Triggers \\ (train/dev/ \\ test)
    \end{tabular} \\
    \hline
    1788 & 36 & 6290/1992/1921 & 2446/830/780\\
    \hline
    \end{tabular}
\end{center}
\caption{\textbf{Dataset details} for DialogRE. With only 2446 labeled triggers in the training set, \drex\;models learn using only a policy gradient and no direct supervision on the remaining 3844 triples.}
\label{tab:DialogRE}
\end{table}
\paragraph{DialogRE Dataset}

We evaluate our models on the DialogRE English V2 dataset\footnote{Dataset collected from https://dataset.org/dialogre/ for research purposes only} which contains dialogues from the Friends TV show \cite{yu-etal-2020-dialogue}, details of which are in Table \ref{tab:DialogRE}. \drex\;models are trained with trigger supervision on less than 40\% of the training data, and make no use of dev or test set triggers. The learning signal for the remaining triples comes entirely from our rewards through a policy gradient.

\paragraph{Evaluation Metrics}
We adopt separate evaluations for relation and explanation extraction.

First, for relation extraction, we evaluate our models using F1 score, following \citet{yu-etal-2020-dialogue}, and additionally calculate the mean reciprocal rank (MRR), which provides further insight into a model's performance. For example, MRR is able to differentiate between a ground truth relation ranked 2nd or 10th, while the F1 score does not. In the dialogRE dataset, multiple relations may hold between a single pair of entities, so we use a variation of MRR which considers all ground truth relations, rather than just the highest-ranked ground truth relation.

For explanation extraction, we focus mainly on manual evaluations, but also propose the Leave-One-Out metric, introduced in section \ref{ablation_study} for an ablation study.

\subsection{Relation Extraction (RE) Evaluation}
In Table \ref{table:relationextraction}, we compare the baseline RE model \rank\textsubscript{BERT} with the methods presented in this paper. We also compare with three other methods which use similarly sized language models, but additionally utilize graph neural networks (GNN): GDPNet\cite{Xue2021GDPNetRL}, TUCORE-GCN\textsubscript{BERT}\cite{lee-choi-2021-graph}, and SocAoG\cite{Qiu2021SocAoGIG}.

First, we see that even though \drex\;is designed to introduce human-understandable explanations, it still has modest improvements over \rank\textsubscript{BERT}, which focuses on RE, while \joint\;has no significant improvement. Next, we see a five point absolute improvement in F1 from the baseline model when using RoBERTa.
The trend from BERT to RoBERTa is similar to results found by \citet{lee-choi-2021-graph}, where changing from a BERT\textsubscript{base} model to RoBERTa\textsubscript{Large}(not shown here) improved their model performance significantly. Additionally, we see a 3 point improvement from \rank\;to \drex\;when using RoBERTa (compared to 0.7 for BERT), which we believe is due to the better performing ranking model, which allows for \drex\;to rely more on the input explanations. Finally, we see that by using GNNs, and task-specific dialogue representations, all three GNN-based methods can improve over the general BERT-based methods.

\begin{table}[t]
\small
\begin{center}
    \begin{tabular}{l|c|c}
        \multicolumn{1}{c|}{\textbf{Model}} & 
        \textbf{F1}($\sigma$) & \textbf{MRR}($\sigma$) \\
        \hline
        \rank\textsubscript{BERT} & 59.2(1.9) & 74.8(1.3)\\
        \joint\textsubscript{BERT} & 59.4(1.7) & 74.0(0.9)\\
        \drex\textsubscript{BERT} & \textbf{59.9}(0.5) & \textbf{75.4}(0.1)\\
        \hline
        \rank\textsubscript{RoBERTa} & 64.2(1.6) & 77.9(1.0)\\
        \joint\textsubscript{RoBERta} & 65.2(0.3) & 78.3(0.3)\\
        \drex\textsubscript{RoBERTa} & \textbf{67.2}(0.3) & \textbf{79.4}(0.3)\\
        \hline
        *GDPNet & 60.2(1.0) & - \\
        *TUCORE-GCN\textsubscript{BERT} & 65.5(0.4) & - \\
        $^{\dagger}$SocAoG & \textbf{69.1}(0.5) & - \\
        \hline
    \end{tabular}
\caption{\textbf{Relation extraction results on DialogRE V2}. \rank\;models are described in Section \ref{RE}, \joint\;models in \ref{sec:joint}, and \drex\;models in \ref{sec:drex}. \rank\textsubscript{BERT} is a replication of BERT\textsubscript{S} from \citet{yu-etal-2020-dialogue}. "*" denotes results taken from \citet{lee-choi-2021-graph} and "$^{\dagger}$" from \citet{Qiu2021SocAoGIG}}
\label{table:relationextraction}
\end{center}
\end{table}

\subsection{Explanation Extraction (EE) Evaluation}
\paragraph{Automatic Evaluation}Although the aim of this paper is not trigger prediction, for completeness and reproducibility, we include results on the test set of triggers in Appendix \ref{triggerprediction}.

\paragraph{Human Evaluation}
To better understand how our model performs in extracting explanations and what challenges still exist, we perform two analyses; a comparative and an absolute analysis. We consider two sets of data for evaluation: samples for the DialogRE test set where \textbf{N}o \textbf{L}abeled trigger exists (\emph{\textbf{NL}}) and samples where the predicted explanation \textbf{D}iffers from the \textbf{L}abeled trigger (\emph{\textbf{DL}}).

\subsubsection{Comparative Analysis}
In Table \ref{table:comparative}, we show the results for pairwise comparisons of explanations predicted by \drex\textsubscript{RoBERTa}\;against 3 baselines: random strings of 1-4 words, predictions from \joint\textsubscript{RoBERTa}, and labeled triggers. For each comparison, we employ 3 crowd-workers\footnote{Amazon Mechanical Turk workers were paid \$0.35 per HIT, where a HIT includes 3 comparisons. We estimate an average HIT completion time of \textasciitilde1.5 minutes, averaging \textasciitilde\$14 per hour. We only accept workers from AUS, CA, and USA.}, who  were given the full dialogue, a natural language statement corresponding to a relational triple, and the two proposed explanations highlighted in the dialogue\footnote{Example HIT included in Appendix \ref{fig:annotation_framework}}. The crowd-workers were asked to specify which of the highlighted explanations was most indicative of the relation, or they could be equal. For each comparison we use a majority vote, and if there was a three-way tie we consider the explanations to be equal. We compare \drex\;with random strings and the joint model on 174 samples from \emph{\textbf{NL}}, as well as 174 samples from \emph{\textbf{DL}}.

In Table \ref{table:comparative} we see that for \emph{\textbf{NL}}, \drex\;produces explanations which were 4.2 times more likely to be outright preferred by crowd-workers than the joint model, suggesting that our reward functions properly guided the explanation policy to learn meaningful explanations on unlabeled data. Surprisingly, we found that on over 12\% of samples with labeled triggers, evaluators outright preferred \drex\;explanations over the ground truth trigger, suggesting that \drex\;indeed finds some explanations which are better than the ground truth trigger.

In Appendix \ref{explanation_comparison}, we include 2 examples comparing explanations from \drex\;and \joint.

\begin{table}[t]
    \small
    \centering
    \begin{tabular}{c|c|c|c}
        \textbf{\drex\textsubscript{RoBERTa}} vs. & \textbf{Win}(\%) & \textbf{Tie}(\%) & \textbf{Lose}(\%) \\
        \hline
        Random (\emph{NL}) & 79.9 & 10.4 & 9.8\\
        \hline
        \joint\textsubscript{RoBERTa} (\emph{NL})& 38.5 & 52.3 & 9.2\\
        \hline
        Ground truth (\emph{DL})& 12.1 & 44.3 & 43.7\\
        \hline
    \end{tabular}
    \caption{\textbf{Human evaluator preferences on explanation extraction methods}. \emph{NL} and \emph{DL} are samples where \textbf{N}o \textbf{L}abeled trigger exists, and where the predicted explanation \textbf{D}iffers from the \textbf{L}abel, respectively. Results presented are percentages of preference.}
    \label{table:comparative}
\end{table}
\begin{table}[t]
    \small
    \centering
    \begin{tabular}{c|c|c|c|c}
         & \begin{tabular}{c}
              Not \\ Indic-\\ative
         \end{tabular} &
         \begin{tabular}{c}
              Incorrect\\Entity\\Pair
         \end{tabular} &
         \begin{tabular}{c}
             Incorrect\\Relation
         \end{tabular} &
         \begin{tabular}{c}
         Indic-\\ative\end{tabular}\\
         \hline
         \emph{NL} & 29 & 19 & 18 & 34\\
         \hline
         \emph{DL} & 19&13&7&61\\
         \hline
    \end{tabular}
    \caption{\textbf{Explanation error analysis} on 100 samples where \textbf{N}o \textbf{L}abeled trigger exists (\emph{NL}) and 100 where the predicted explanation \textbf{D}iffers from the \textbf{L}abel (\emph{DL}).}
    \label{table:absolute}
\end{table}

\subsubsection{Absolute Analysis}
\label{abs_analysis}
To better understand the quality of \drex's explanations, we randomly sample 100 from both \emph{\textbf{NL}} and \emph{\textbf{DL}} for a fine-grained analysis. We classify the explanations into 4 categories: not indicative, incorrect entity-pair, incorrect relation, and indicative. "Indicative" and "Not indicative" have the obvious meanings, "Incorrect entity-pair" means that an explanation actually explains the correct relation, but between the incorrect entity-pair, and "Incorrect relation" means that the explanation indicates a relation different from the desired relation.

Table \ref{table:absolute} shows the results. Interestingly, we see in the \emph{\textbf{NL}} set, that errors were equally likely to come from either an explanation indicating the relation for an incorrect entity-pair as for the incorrect relation altogether. This is in contrast to the \emph{\textbf{DL}} set, where \drex\;was nearly half as likely to predict an explanation for an incorrect relation as it was for an incorrect entity-pair. 

Additionally, in our fine-grained analysis, we also considered whether a relational triple was identifiable from the context alone and found that nearly 20\% of the 200 samples had ambiguities which could not be resolved without outside knowledge. This suggests that there is likely a maximum achievable relation extraction score on the DialogRE dataset under the current setting.


\begin{table}[t]
    \small
    \centering
    \begin{tabular}{l|c|c}
        \multicolumn{1}{c|}{\textbf{Model}} & \textbf{F1} & \textbf{Leave-one-out}($\downarrow$)  \\
        \hline
        \drex\textsubscript{RoBERTa} (Full) & 67.2 & 83.9\\
        \hline
        \quad - reranking reward & 66.0 & 84.9\\
        \quad - LOO reward & 67.1 & 85.4\\
        \hline
    \end{tabular}
    \caption{\textbf{Ablation study} on reward functions. Leave-One-Out metric (LOO) measures how salient a predicted explanation is in determining a relation and is further defined and motivated in $\S$\ref{ablation_study}. Smaller LOO is better.}
    \label{table:ablation}
\end{table}
\subsection{Ablation Study}
\label{ablation_study}
To assess the benefit of each proposed reward individually, we perform an ablation study on the reward functions. In order to study explanation quality automatically, we introduce a new metric for explanation quality; the Leave-One-Out metric.

\begin{figure*}[t]
    \small
    \centering
    \includegraphics[width=\textwidth]{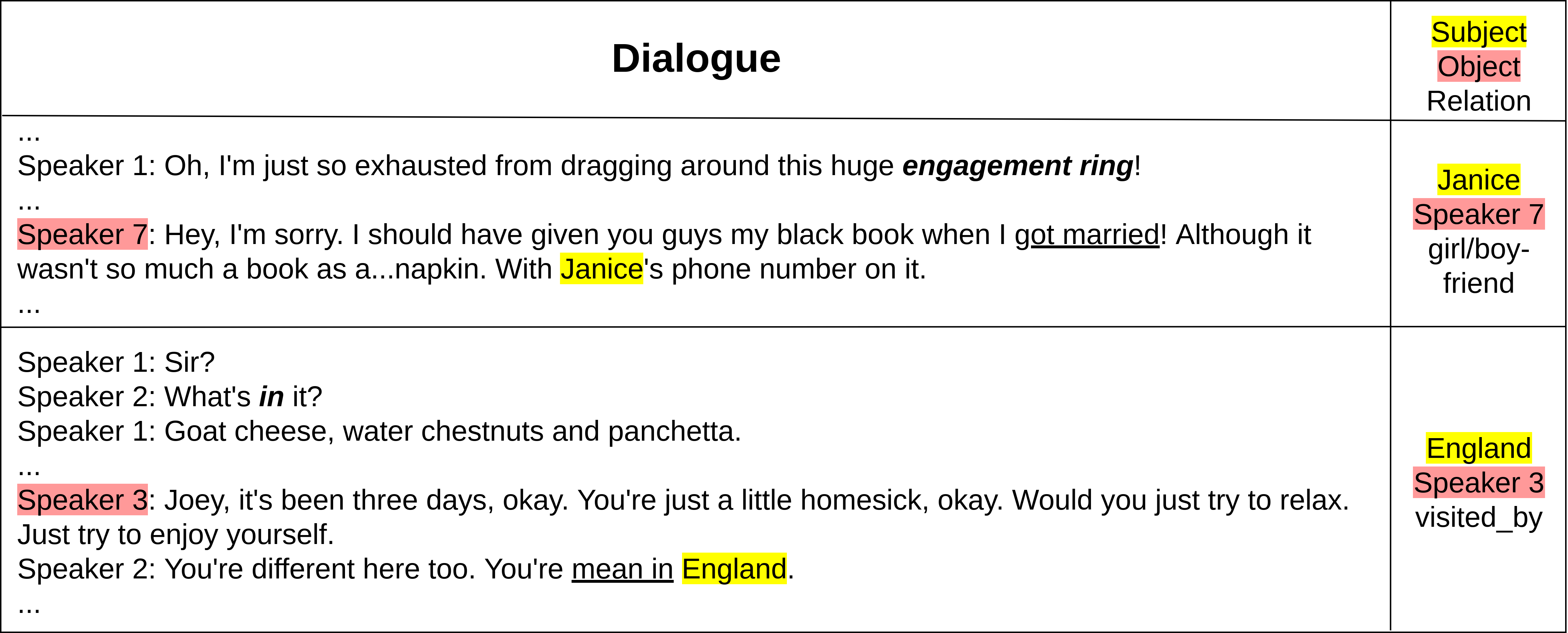}
    \caption{Two examples comparing predicted explanations from \drex\;(\underline{underlined}) and \joint\;(\emph{\textbf{bold}}).}
    \label{fig:drex_joint}
\end{figure*}

The Leave-One-Out (LOO) metric has a theoretical basis in the works of \citet{DBLP:journals/corr/LiMJ16a} and \citet{10.1145/2939672.2939778}, where \citet{DBLP:journals/corr/LiMJ16a} use word erasure to determine a "word importance score". Here we define LOO formally. For a relation extraction model \rank, an explanation extraction model \expl, and a dataset $\mathcal{D}$, LOO is calculated as\[
LOO(\rank, \expl, \mathcal{D}) = \frac{\text{F1}_{\rank}(\mathcal{D_{\text{MASK}}(\expl)})}{\text{F1}_{\rank}(\mathcal{D})}
\]
where F1\textsubscript{\rank}($\mathcal{D}$) is the F1 score of \rank\;on $\mathcal{D}$ and $\mathcal{D}$\textsubscript{MASK}(\expl) is the dataset where explanations predicted by \expl\;are replaced by mask tokens. The LOO metric calculates how essential the predicted explanations are to the ability of the relation extraction model.

To show that LOO is an appropriate measure of explanation quality, we compute the Pearson correlation coefficient between token F1 score and LOO scores for models on labeled triggers, found in Table \ref{table:triggerprediction}. With 6 models trained on 5 random seeds each, we have 30 data points and a correlation coefficient of $-87.4$ with $p=2.4*10^{-8}$. Because we calculate the coefficient with respect to human-annotated triggers, this suggests that a low LOO correlates with explanations that humans would determine as indicative of the given relation.

For our experiments, we always calculate LOO using the baseline model, \rank\textsubscript{BERT}. From the results in Table \ref{table:ablation}, we see that both reward functions benefit the final results. Compared with \rank\textsubscript{RoBERTa}, \drex\textsubscript{RoBERTa} gains 3 F1 points, but without the reranking reward, the model only gains 1.8 F1 score or 60\% of the total possible improvement. This performance loss demonstrates that the reranking reward is critical to attaining the best score in relation extraction. Similarly, without the leave-one-out reward, the model's explanation quality, measured in LOO, is 1.5 points, or nearly 10\% worse, demonstrating that the leave-one-out reward is beneficial in guiding the model to salient explanations.

\subsection{Explanation Samples}
\label{explanation_comparison}
Figure \ref{fig:drex_joint} shows two samples comparing explanations from \drex\;and \joint. In both examples, even though there was no labelled trigger, each model was able to predict an explanation which correlates with the relation. Specifically, "engagement ring" and "got married" are related to the girl/boyfriend relation, and "in" and "mean in" can be associated with the visited\_by relation. However, the bottom example shows that \joint\;did not consider the context surrounding it's explanation. The conversation is about food, and the visited\_by relation is not relevant. On the other hand, \drex\;finds the phrase "you're mean in", where "you're" refers to speaker3, and "in" refers to "England". This is clearly an explanation which indicates the correct relation between the correct entities.

\subsection{Reduced Labels}
\label{negative_results}
All previous results use 100\% of labeled triggers in the DialogRE dataset, which covers 40\% of all relational triples. To test how few labeled triggers \expl\;requires in order to learn meaningful explanations we ran a small scale experiment (1 random seed) using labeled triggers from only 5, 10, and 20\% of relational triples. However, in the small tests we ran, we found that at 20\% labeled triggers the \expl\;model mostly predicts no explanations. Furthermore, at 10\% and fewer labeled triggers, the model converges to the trivial solution in the explanation space which is to never predict any tokens.

We believe that this issue is due, in part, to two challenges: the search space over all possible start/end tokens is too large, and the policy gradient has a high variance.
Although these results may seem discouraging, we believe this challenge can be overcome in the future by using algorithms which reduce variance in the policy gradient and by initializing \expl\;with a model pre-trained in span extraction.

\section{Limitations and Future Work}

Firstly, this study focuses on learning explanations as well as relations in dialogue and DialogRE is the only currently available dataset with annotations for both tasks. A limitation of this study is the small scale at which we were able to test the methods. A future direction would be to learn explanations on a different RE dataset and use the pre-trained model in \drex, however it would be non-trivial for a model to transfer explanations learned on a wildly different domain. Additionally, it is theoretically possible to train \drex\;with no labeled triggers at all, however, we were unsuccessful and in Section \ref{negative_results} we discuss these and additional negative results.

This study focuses on relations and entities found in multi-party conversations, and while there are similarities between the dialogue domain, medical literature, and wikipedia (e.g., multi-entity, multi-relation), it is not clear whether the methods from this paper can transfer to other such domains. We plan to investigate how well the proposed methods transfer to relations and entities in other domains such as news and web text \cite{zhang-etal-2017-position} and for other types of semantic relations as in \citet{hendrickx-etal-2010-semeval} or \citet{yao2019docred}.

We acknowledge that this study is English-focused, and it is not clear that these methods can transfer to languages in other families such as afro-asiatic or sino-tibetan. Additionally, we think that it would be very interesting to see how these methods perform on languages with very different linguistic features; for example, languages with inflection such as Finnish. We leave non-English and multilingual variations of these methods to future work.

In this work, we do not focus on improving state-of-the-art trigger prediction. However, we recognize that trigger annotation is labor-intensive, and a possible use of \drex\;would be to use predicted labels as a form of weak supervision for a system whose goal \emph{is} to improve on trigger prediction.

Finally, while relation extraction and relation explanations are an obvious pair of candidate tasks for the proposed methods, the methods are general, and may be useful for other related task pairs. For example, \citet{Albalak2022FETAAB} introduce a dataset with 132 possible task-pairs, all with limited overlapping annotations.
\section{Related Work}
Recently, there have been numerous information extraction tasks proposed which involve dialogues, including character identification \cite{zhou-choi-2018-exist}, visual coreference resolution \cite{yu-etal-2019-see}, emotion detection \cite{zahiri2018emotion}.

New settings for relation extraction have also been proposed, such as web text \cite{DBLP:journals/corr/abs-2102-09681} and, in many ways similar to dialogue, document text \cite{yao2019docred}. 
There have also been methods developed to include explanations in similar natural language understanding tasks \cite{NEURIPS2018_4c7a167b,kumar-talukdar-2020-nile,liu-etal-2019-towards-explainable,lei-etal-2016-rationalizing}. There have even been methods developed which, similarly to our re-ranking, make use of an explanation as additional information \cite{hancock-etal-2018-training}.

The work by \citeauthor{shahbazi-etal-2020-relation} is aligned with our study. They also focus on relation extraction with explanations; however, their method is based on distant supervision from bags of sentences containing an entity-pair. Due to the cross-sentence nature of relations in dialogue, their method is not applicable here, although we draw inspiration from their work. They explain their model by considering the salience of a sentence to their model's prediction, similarly to our leave-one-out reward.

Also relevant to our work is that by \citeauthor{bronstein-etal-2015-seed}. Their work focuses on the task of semi-supervised event trigger labeling, which is very similar to our semi-supervised prediction of relation explanations. In their work, they use only a small seed set of triggers and use a similarity-based classifier to label triggers for unseen event types.

Finally, there have been multiple recent works in dialogue RE which perform quite well by using graph neural networks \cite{Xue2021GDPNetRL,Qiu2021SocAoGIG,lee-choi-2021-graph}. However, they focus only on RE and not on explaining the relations.
\section{Conclusion}
In this work, we demonstrated that not only is it possible to extract relation explanations from multi-party dialogues, but these explanations can in turn be used to improve a relation extraction model. We formulated purpose-driven reward functions for training the explanation model and demonstrated their importance in learning high quality explanations. Our proposed approach, \drex, is powered by a very simple reformulation of the traditional relation extraction task into a re-ranking task.

\section{Ethical and Social Considerations}
The methods proposed in this work on their own are not nefarious, however, proposed explanations should not be blindly accepted as fact. For the same reasons that language models may have ethical and social risks, so may our algorithm which is built on top of such language models. While we test only on TV show dialogues, were this technology to be put to use in non-scripted, real life conversations, there would need to be very thorough analysis of any ethical risks that the proposed explanations may have.

\bibliography{anthology,d_rex}
\bibliographystyle{acl_natbib}

\newpage

\appendix
\section{Trigger prediction}
\label{triggerprediction}
\begin{table}[t]
    \small
    \centering
    \begin{tabular}{l|c|c|c}
         \multicolumn{1}{c|}{\textbf{model}} & \textbf{token F1}($\sigma$) & \textbf{EM}($\sigma$) & \textbf{LOO}($\sigma$)\\
         \hline
         \expl\textsubscript{BERT}& 62.1(3.1) & 54.1(1.9) & 82.2(0.4)\\
         \joint\textsubscript{BERT} & 43(1.3) & 38.6(1.4) & 89.0(1.0)\\
         \drex\textsubscript{BERT} & 50.5(1.1) & 45.7(1.7) & 84.4(1.6)\\
         \hline
         \expl\textsubscript{RoBERTa} & 66.5(2.2)& 58.4(2.0) & 82.2(0.4)\\
         \joint\textsubscript{RoBERTa} & 49(0.7) & 47(0.7) & 86.2(0.8)\\
         \drex\textsubscript{RoBERTa} & 57.2(2.1) & 51.6(1.6) & 83.9(0.4))\\
    \end{tabular}
    \caption{\textbf{Trigger prediction results}. Leave-One-Out metric (LOO) measures how salient a predicted explanation is in determining a relation and is further defined in $\S$\ref{ablation_study}. Smaller LOO is better.}
    \label{table:triggerprediction}
\end{table}

In Table \ref{table:triggerprediction}, we compare our methods for supervised explanation extraction with \drex. Interestingly, we find that the joint model achieves the lowest F1 score for both the BERT and RoBERTa models. \joint\textsubscript{BERT} scores nearly 20 points below its counterpart BERT model, while the \joint\textsubscript{RoBERTa} model cuts that difference to just over 15 points below its RoBERTa counterpart. On the other hand, \drex\;maintains a token F1 score within 10 points of its counterpart even though it has been trained to generalize beyond the labeled triggers.

\section{Hyperparameters}
\label{hyperparameters}
All models are trained using the AdamW optimizer \cite{DBLP:journals/corr/abs-1711-05101} with a learning rate of $3e$-$5$ and batch sizes of $30$. To determine the best learning rate, \rank\;and \expl\;models were trained using learning rates in \{$3e$-$6$, $1e$-$5$, $3e$-$5$, $1e$-$4$\}. The best learning rate, $3e$-$5$, was determined by performance on a held out validation dataset. Baseline models (\rank, \expl, and \joint) are trained for at most $30$ epochs and we use validation-based early stopping to determine which model to test. \drex\;models are trained for at most $30$ additional epochs with the best model determined based on relation extraction F1 scores computed on validation data. We found the best validation result to always occur within the first 30 epochs. All experiments were repeated five times and we report the mean score along with standard deviation.  To train the joint model, we do not calculate $\mathcal{L}_{EX}$ for relational triples which do not have a labeled trigger and we select $\alpha$ from \{0.25,0.5,0.75\} and set $\alpha$ to 0.5 based on validation performance.

\section{Crowd-Worker Sample}
\begin{figure*}[t]
    \centering
    \includegraphics[scale=0.6]{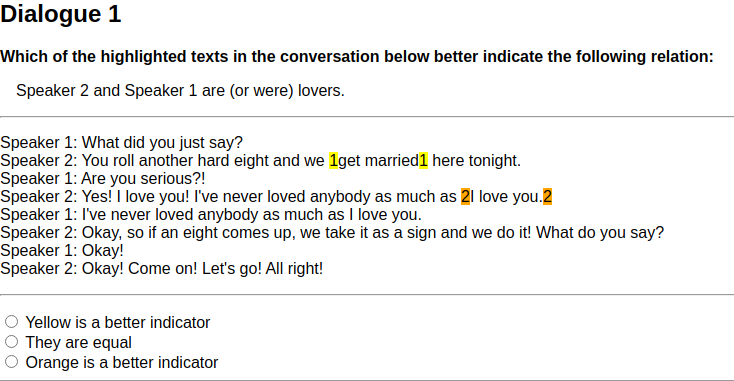}
    \caption{A sample HIT that was presented to crowd-workers for the comparative study of explanations.}
    \label{fig:annotation_framework}
\end{figure*}

In Figure \ref{fig:annotation_framework}, we show a sample HIT that was provided to crowd-workers. Each crowd-worker was shown three examples. The layour is as follows: the top always asks the worker to decide which of the highlighted texts is a better indication of the relation. Next, a natural language interpretation of the relational triple is given, in this case, "Speaker 2 and Speaker 1 are (or were) lovers". Then, we show the entire dialogue along with highlighted spans of text for each explanation. Finally, at the bottom, we always provide the user with three choices: yellow is better, equal, or orange is better, where the user is only allowed to select one option.

\end{document}